\def\year{2019}\relax
\documentclass[letterpaper]{article} 
\usepackage{aaai19}  
\usepackage{times}  
\usepackage{helvet}  
\usepackage{courier}  
\usepackage{url}  
\usepackage{graphicx}  

\usepackage{multirow}
\usepackage[ruled]{algorithm2e}
\usepackage{mathrsfs}

\usepackage{times}
\usepackage{latexsym}

\usepackage{url}

\usepackage{color}
\usepackage{xcolor}
\usepackage{booktabs} 

\usepackage{amsmath}
\usepackage{amsfonts,amssymb}

\usepackage{verbatim}
\usepackage{bm}
\usepackage{sansmath}
\usepackage{balance}
\usepackage{longtable}
\usepackage{epstopdf}

\usepackage{epsfig}
\usepackage{subfigure}

\newcommand{\vpara}[1]{\vspace{0.01in}\noindent\textbf{#1 }}
\newcommand{\hide}[1]{} 

\usepackage{mathrsfs}
\begin{document}
%
\title{Relation Mention Extraction from Noisy Data with Hierarchical Reinforcement Learning}

\author{Jun Feng$^\ddag$, Minlie Huang$^\S$\thanks{Corresponding author: Minlie Huang, aihuang@tsinghua.edu .cn}, Yijie Zhang$^\S$, Yang Yang$^\dag$, and Xiaoyan Zhu$^\S$ \\
	$^\ddag$ State Grid  Corporation of China\\
	$^\S$ State Key Lab. of Intelligent Technology and Systems, National Lab. for Information Science and Technology\\
	Dept. of Computer Science and Technology, Tsinghua University, Beijing 100084, PR China \\
	$^\dag$ College of Computer Science and Technology, Zhejiang University\\
	{\tt JuneFeng.81@gmail.com} ,\quad {\tt aihuang@tsinghua.edu.cn}, \quad{\tt yj$\_$zhang15@mails.tssinghua.edu.cn} \\ 
	{\tt yangya@zju.edu.cn} ,\quad {\tt zxy-dcs@tsinghua.edu.cn}}

\maketitle
\begin{abstract}
In this paper we address a task of {\it relation mention extraction} from noisy data: extracting representative phrases for a particular relation from noisy sentences that are collected via distant supervision.
Despite its significance and value in many downstream applications, 
this task is less studied on noisy data.
The major challenges exists in 1) the lack of annotation on mention phrases, and more severely, 2) handling noisy sentences which do not express a relation at all.
To address the two challenges, we formulate the task as a semi-Markov decision process and propose a novel hierarchical reinforcement learning model. Our model consists of a top-level sentence selector to remove noisy sentences, a low-level mention extractor to extract relation mentions, and a reward estimator to provide signals to guide data denoising and mention extraction without explicit annotations.  		
Experimental results show that our model is effective to extract relation mentions from noisy data. 
\end{abstract}

	\section{Introduction}
\begin{figure}[t]
		\centering
		\includegraphics[width=0.45\textwidth]{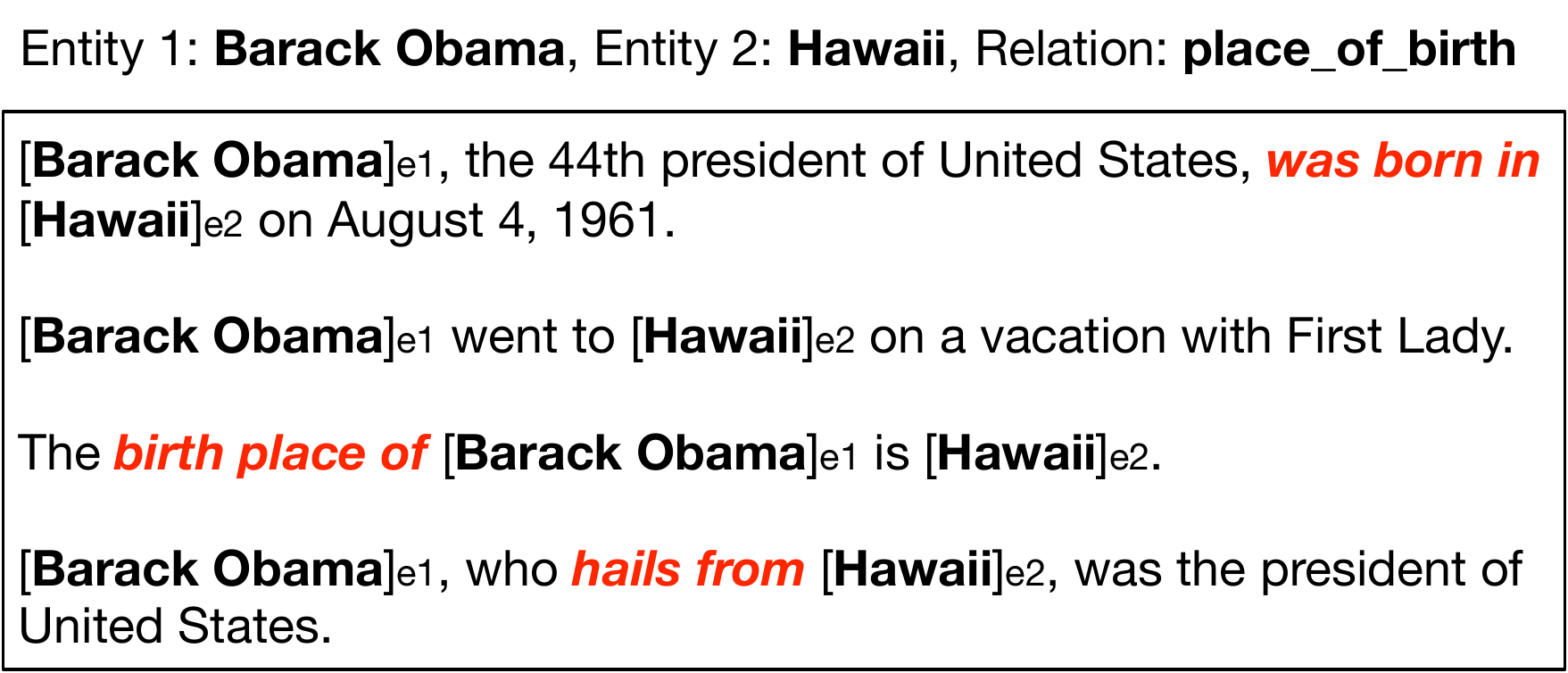}
		\caption{Illustration of relation mention extraction from noisy sentences. Words in red are relation mentions. 
		}
		\label{fig:mention}
	\end{figure}

	%
	The increasing demand for structured knowledge 
	has significantly advanced the research of named entity recognition and relation extraction. 
	Extensive prior research has studied extracting  entities~\cite{borthwick1998exploiting,chiu2016named,xu2017local} and relations~\cite{bunescu2005shortest,mintz2009distant,zeng2014relation,zheng2017joint} from a plain text. 
	Figure~\ref{fig:mention} illustrates an example of relation extraction, where the relation `` place\_of\_birth'' between two entities ``Barack\_Obama'' and ``Hawaii'' is detected since the expression ``{\it was born in}'' suggests the relation ``place\_of\_birth'' directly. Such representative expressions are referred to as \textit{relation mention}.

	Relation mentions can be valuable resources in many downstream tasks and benefit many applications such as relation extraction,  question answering, and language inference. Moreover, it offers good interpretability to reveal the textual evidence for a detected relation, and further, we can study the language variety in relation mention: there are various phrases and ways to express the same relation. For instance, for the ``{ place\_of\_birth}'' relation, there are many expressions such as {\it ``the birth place'', ``was born in'', ``hails from''}, and so on.
	
    Relation mention extraction in this paper is defined as follows: given a relation $r$, and a set of sentences containing an entity pair and associated with a {\bf noisy} relation label $r$\footnote{The relation label is automatically generated under the distant supervision assumption. \textit{Noisy} means that some sentences may not mention the automatically-labeled relation $r$ at all.}, the task is to extract a set of representative phrases for relation $r$ (e.g., ``place\_of\_birth''), such as ``{\it the birth place}'', ``{\it hails from}'', and ``{\it was born in}''. We term the task as {\it relation mention extraction}.

	Many existing studies only focus on sentence-level relation classification that predicts whether a sentence mentions a relation \cite{riedel2010modeling,hoffmann2011knowledge,li2014incremental,miwa-bansal:2016:P16-1,ren2017cotype,zheng2017joint}. However, they do not concern the words or phrases that describe a relation.
	Our problem also differs from Open IE ~\cite{banko2007open,fader2011identifying,angeli2015leveraging}, in that such systems do not need to normalize different expressions (e.g., ``{\it the birth place}'' and ``{\it was born in}'') to the same canonical relation (e.g., ``place\_of\_birth''),  as shown in Figure~\ref{fig:mention}. 
	Some works deal with the noisy labeling issue on relation label ~\cite{takamatsu2012reducing,zeng2015distant,feng2018reinforcement}, but they do not involve  relation mention extraction.

	
	
	There are two major challenges for relation mention extraction. \textbf{First}, the sentences for a relation are constructed by distant supervision \cite{mintz2009distant,zeng2015distant}, and are hence noisy where a sentence may not describe the relation at all. Extraction from noisy sentences will definitely lead to undesired, incorrect relation mentions. 
	\textbf{Second}, it is too costly to conduct mention annotation to specify which words or phrases mention a relation in a sentence, particularly in the setting of large-scale relation mention extraction. Instead, there is only a very weak signal available, indicating that a sentence (noisy itself) {\it might} describe a relation. 
	
	To address these challenges, we devise a hierarchical reinforcement learning~\cite{sutton1999between} model to address the task of relation mention extraction from noisy sentences. The model consists of three components: a top-level sentence selector for selecting correctly labeled sentences that express a particular relation, a low-level mention extractor for identifying mention words in a selected sentence, and a reward estimator for providing signals to guide sentence denoising and mention extraction without explicit annotations. The intuition behind this model is as follows: if a high-quality sentence is selected, it will facilitate relation mention extraction, and in return, the extraction performance will signify the fitness of sentence selection.  
	
	Our model works as follows:
	at the top level, the agent decides whether a sentence should be selected or not from a sentence bag\footnote{A sentence bag contains sentences labeled as the same relation}; once the agent selects a sentence, it enters into a low-level RL process for mention extraction. When the low-level process completes its task, the agent will return back to the top-level process and continues to tackle the next sentence in the bag. Since we have no explicit annotations on either sentence (whether a sentence truly describes a relation) or word (which words are a relation mention), the problem can be formulated as a natural sequential decision problem and the policy learning in the high-level and low-level processes is guided by the delayed rewards (the likelihood of relation classification), which is a weak, indirect supervision signal for policy learning.  
	

	Our contributions are as follows: 
	\begin{itemize}
		\item We study the task of relation mention extraction in new settings: from noisy sentences and with only  weak supervision, that is, there is no explicit annotations on sentences or mention words.  
		
		\item We propose a novel hierarchical reinforcement learning model which consists of a top-level sentence selector for removing noisy sentences, a low-level extractor for extracting relation mentions, and a reward estimator for offering supervision signals to guide data denoising and mention extraction.
	\end{itemize}
	
	\section{Related Work}
	\label{sec:related}
	We deal with relation mention extraction in this paper. 
	As closely related tasks, named entity recognition (NER) and relation extraction (RE) have attracted considerable research efforts recently. 
	NER locates entity's mentions in a plain text~\cite{borthwick1998exploiting,chiu2016named,xu2017local,katiyar2017going}. 
	As entity mentions are less diverse and it is easier to access high-quality labels for NER, this task is usually formulated as a full supervision problem (e.g., sequential labeling). 
	The goal of RE is to extract semantic relations between two given entities. 
	Many researchers have explored models based on handcrafted features~\cite{mooney2005subsequence,guodong2005exploring} or deep neural networks~\cite{socher2012semantic,zeng2014relation,DBLP:conf/acl/SantosXZ15,lin2017neural}. 

	The most relevant to our work is Open IE~\cite{banko2007open,wu2010open,hoffmann2011knowledge,angeli2015leveraging}, which extracts 
	triples that contain
	two entities and a relation mention.
	However, there is no need to normalize different expressions to a canonical relation in Open IE systems. 
	
	There exists a large amount of studies for sentence-level relation classification which predicts whether a sentence describes a relation but without specifying a token span as mention~\cite{riedel2010modeling,hoffmann2011knowledge,li2014incremental,miwa-bansal:2016:P16-1,ren2017cotype,zheng2017joint}.
	\cite{wang2016relation} and \cite{huang2016attention} adopted attention mechanisms to highlight some words in a sentence as the clues of a relation. 
	However, such methods can only detect separate words but do not consider the dependency between words. 
	
	There are also some works~\cite{feng2018reinforcement,zeng2018large} using reinforcement learning for relation extraction from noisy data. However, they target more on relation classification instead of mention extraction.
	%
	Our work is inspired by \cite{feng2018reinforcement} where an instance selector was used to remove noisy sentences. However, the supervision signal for sentence selection is sparse as there is only a delayed reward available after all selection in a bag is completed. By contrast, our model is more straightforward: the top-level sentence selector can receive an intermediate reward after each selection from the low-level mention extractor and obtains direct feedbacks to guide policy learning.
	

	\begin{figure}
		\centering
		\includegraphics[width=0.5\textwidth]{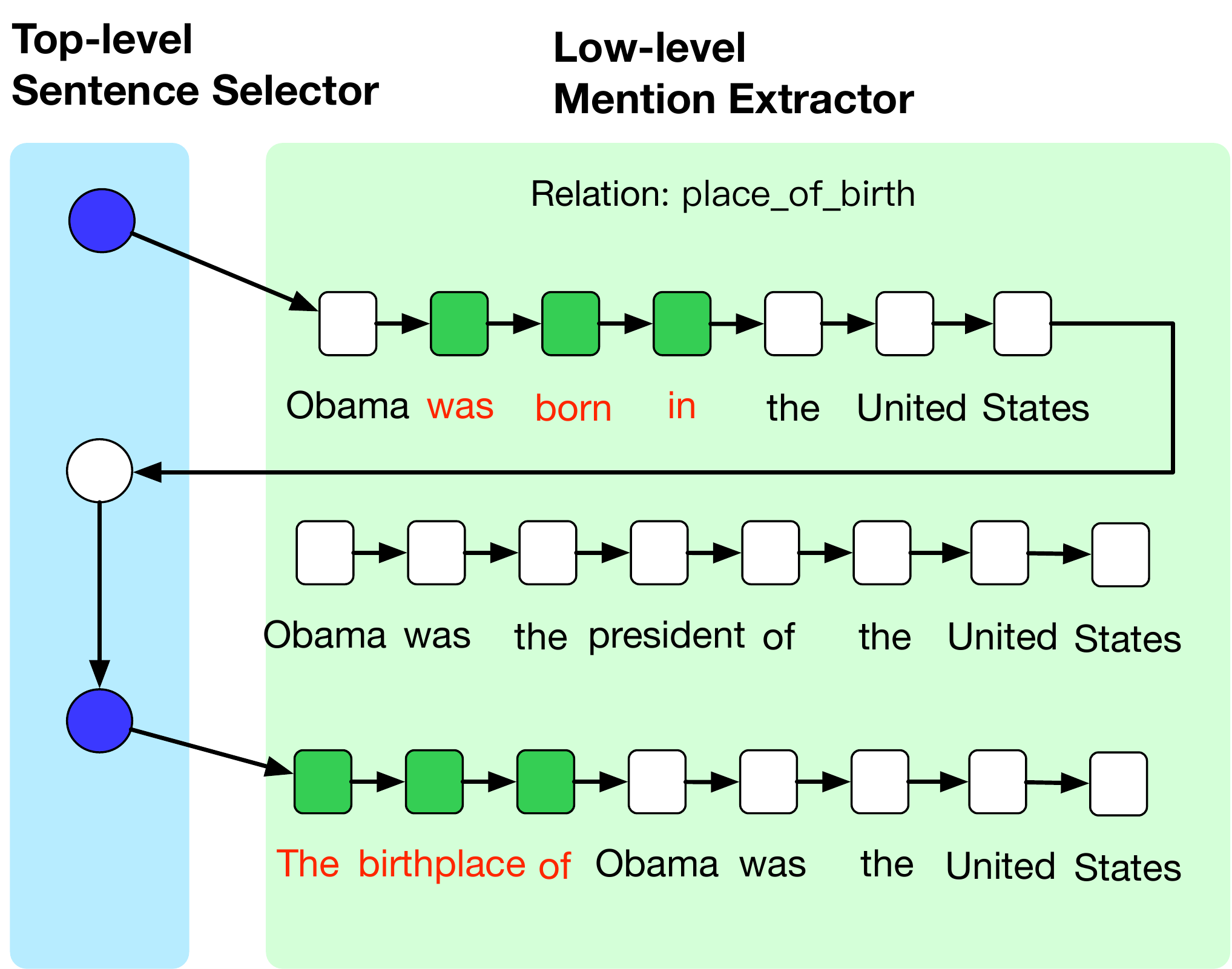}
		\caption{The hierarchical decision making process to extract mentions for relation \textit{``place\_of\_birth''}. Blue circles denote selected sentences (white for unselected sentences), and green squares indicate mention words (white squares means non-mention words). Words in red are mention words.
		}
		\label{fig:hrl}
	\end{figure}	
	
	\section{Methodology}
	\label{sec:model}

	\subsection{Problem Definition}
	We formulate the task of relation mention extraction from noisy data as follows: given a relation $r$ and a sequence of $<$sentence, relation$>$ pairs as $X = \{(x_1, r), (x_2, r), \dots, (x_n, r)\}$, the goal is to extract a set of representative phrases for relation $r$. 
	Each $x_i$ is a sentence associated with two entities $(h, t)$ and a \textbf{noisy relation label} $r$, produced by distant supervision \cite{mintz2009distant}. In other words, a sentence $x_i$ may not express relation $r$ at all.
	
	The challenges for relation mention extraction come from: 1) there are noisy relation labels, and 2) there is no word-level mention annotation.

	\subsection{Overview}
	
	As illustrated in Figure~\ref{fig:hrl}, the process of relation mention extraction works as follows: the agent first decides whether a sentence expresses a given relation; if the agent predicts so, it will scan the words in the sentence one by one to identify the mention words; otherwise, the agent directly skip the current sentence. The agent continues to tackle the next sentence until all the sentences for the same entity pairs are handled.
	The above process can be naturally formulated as a semi-Markov decision process. We thus address the task in the framework of hierarchical reinforcement learning ~\cite{sutton1999between,dietterich2000hierarchical}. The hierarchical reinforcement learning process has two tasks: a top-level RL task which takes an {\it option} for data denoising, deciding whether a sentence should be selected; and a low-level RL task that makes {\it primitive actions} for mention extraction, deciding which words are part of a relation mention.
	
\begin{figure}[t]
		\centering
		\includegraphics[width=0.5\textwidth]{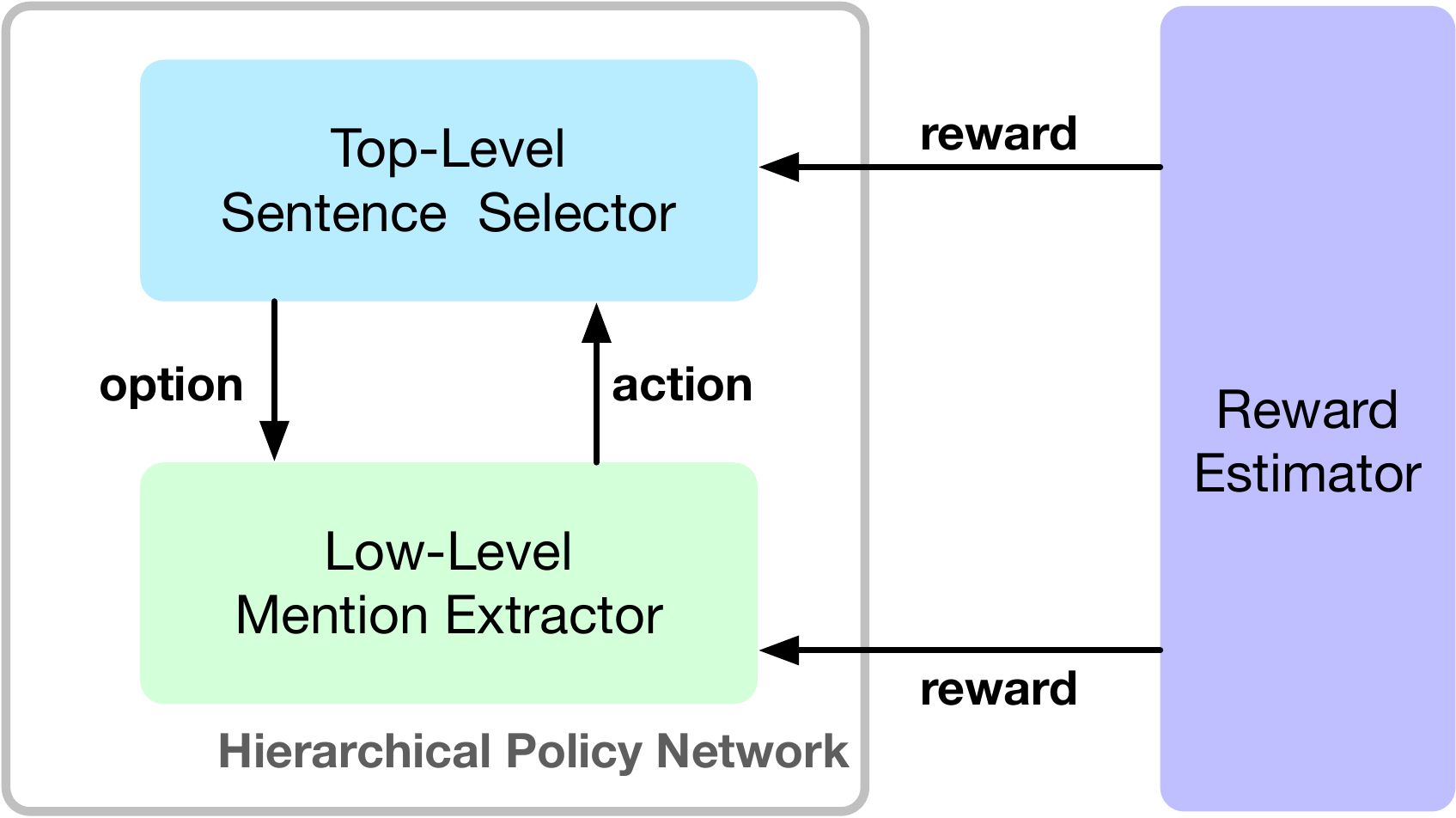}
		\caption{The hierarchical reinforcement learning model.}
		\label{fig:framework}
	\end{figure}

	As shown in Figure~\ref{fig:framework}, our model consists of three components: a top-level sentence selector, a low-level mention extractor, and a reward estimator. 
	The sentence selector scans the sentences in a bag and takes \textit{options} (top-level action) to determine whether a sentence describes a relation. The mention extractor performs a sequential scan on a selected sentence and takes actions on whether a particular word in the sentence is part of a relation mention. 
	As there are no explicit supervision for either the selector or the extractor, we pretrain a relation classifier as the reward estimator to guide the policy learning in the two modules. 
	
	\subsection{Reward Estimator}
	\label{sec:reward}
	We adopt a CNN classifier to offer supervision signals to help estimate the rewards for the sentence selector and the mention extractor. The supervision signal is measured by the likelihood $\mathcal{P}(r|x; \mathbf{\Phi})$ of relation classification for a given sentence $x$.
	Following~\cite{feng2018reinforcement}, the CNN network has an input layer, a convolution layer, a max pooling layer, and a non-linear layer from which the representation is used for relation classification.
	
	
	\noindent \textbf{CNN Structure.}
	The CNN structure can be briefly described as below:
	\begin{equation}
	\mathbf{L} = \text{CNN}(\mathbf{x})
	\end{equation}
	where $\textbf{x}$ is the input vectors and $\mathbf{L} \in \mathbb{R}^{d_s}$ is the result of the max pooling layer.
	In this structure, there is a convolution layer, and a max pooling layer. The convolution operation is performed on 3 consecutive words, and the number of feature maps $d_s$ is set to $230$, the same as \cite{lin2016neural}. Hence, the convolution parameters are $\mathbf{W}_{f} \in \mathbb{R}^{d_s \times (3d)}$ and $\mathbf{b}_f \in \mathbb{R}^{d_s}$. 
	
	Then, the relation classifier estimates $\mathcal{P}(r | x; \mathbf{\Phi})$ as follows:
	\begin{equation}
	\mathcal{P}(r|x; \mathbf{\Phi}) = softmax(\mathbf{W}_r*tanh(\bm{L}) + \mathbf{b}_r)
	\label{eq:reward_estimator}
	\end{equation}
	where $\mathbf{W}_r \in \mathbb{R}^{n_r \times d_s}$ and $\mathbf{b}_r \in \mathbb{R}^{n_r}$ are parameters in the fully-connected layer, $n_r$ is the total number of relations, and the parameters $\mathbf{\Phi}=\{\mathbf{W}_f, \mathbf{b}_f, \mathbf{W}_r,\mathbf{b}_r\}$.
	
	This probability $\mathcal{P}(r | x; \mathbf{\Phi})$ is used to estimate the rewards to the sentence selector and the mention extractor, see Eq. \ref{eq:reward_selector} and Eq. \ref{eq:reward_extractor}. 
	
	
	\noindent \textbf{Loss function.}
	Given a training set $X$, cross-entropy is used as loss function to train the CNN classifier:
	\begin{equation}
	\mathcal{J}(\Phi) = -\frac{1}{|X|}\sum_{i = 1}^{|X|} \log \mathcal{P}(r|x_i; \bm{\Phi})
	\label{eq:cnn_loss}
	\end{equation}

	\subsection{Top-Level Sentence Selector}
	The top-level sentence selector aims to select a sentence that truly mentions the given relation. A selected sentence will then be passed to the low-level mention extractor for further mention extraction. As we do not have an explicit supervision for the sentence selector, we measure the utility of the selected sentences as a whole using a final reward. Thus, this RL process terminates when all the sentences are scanned. In what follows, state $s_t^h$, option $g_t^h$ and reward $r_t^h$ at step $t$ (corresponding to the $t$-th sentence) will be introduced.
	
	\noindent \textbf{State.}
	The state $\bm{s}_t^h$ consists of the information about tthe current sentence, the already selected sentences, the relation label, and the extracted relation mentions from the previously selected sentences:
	\\
	1) The vector representation of the current sentence, which is obtained from the non-linear layer of the CNN classifier for relation classification;
	\\
	2) The average of the sentence representations of the chosen sentences;
	\\
	3) The one-hot representation of a given relation;
	\\
	4) The representation of the extracted relation mentions, which is the average of the word vectors of all the mention words.
	
	\noindent \textbf{Option.}
	The option $g_t^h \in \{0, 1\}$ where 1 means the $t$-th sentence is selected. We sample the value of $g_t^h$ from the policy function as follows:
	\begin{equation}
	\mu(g_t^h|s_t^h;\bm{\theta}^h) = \sigma(\bm{W}^h *\bm{s}_t^h + \bm{b}^h)
	\label{eq:option}
	\end{equation}
	
	\noindent where $\sigma(.)$ is the sigmoid function with the parameter $\bm{\theta}^h = \{\bm{W}^h, \bm{b}^h\}$.
	
	\noindent \textbf{Reward.}
	At each step $t$, if the sentence is selected, the sentence selector will receive an intermediate reward $r_t^h$ which is the delayed reward received by the low-level mention extractor on the $t$-th sentence, as defined by Eq. \ref{eq:reward_extractor}; otherwise, the intermediate reward is set as $0$.
	
	In addition to the intermediate rewards, a final reward is computed to measure the utility of all the chosen sentences, when the top-level selector completes its scan on all the sentences for a given relation:
	\begin{equation}
	r_{final}^h = \frac{1}{|\hat{X}|}\sum\limits_{x_j \in \hat{X}} \log \mathcal{P}(r | x_j) 
	\label{eq:reward_selector}
	\end{equation}
	where $\hat{X}$ ($\subseteq X$) contains the selected sentences, and $r$ is the given relation. $\mathcal{P}(r | x_j)$ is provided by the reward estimator, see Eq. \ref{eq:reward_estimator}.
	
	\subsection{Low-Level Mention Extractor}
	Once the top-level sentence selector chooses a sentence $x_t$, the low-level mention extractor will scan sequentially the words in $x_t$ to identify relation mention words given relation $r$. At each step $j$, the mention extractor makes a decision on whether the $j$-th word is part of the relation mention. This low-level RL process terminates after the last word is scanned.
	
	\noindent \textbf{State.}
	The state $s_j^l$ encodes the information about the current words, the already chosen words in the sentence, and the relation:
	\\
	1) The vector representation of the current words;
	\\
	2) The representation of the chosen mention words, which is the average of the word embeddings of all the chosen words;
	\\
	3) The one-hot representation of the relation.
	
	\noindent \textbf{Action.}
	The action $a_j^l \in \{0, 1\}$ where 1 means the $j$-th word is selected as a mention word. We sample $a_j^l$ from the policy function:
	\begin{equation}
	\pi(a_j^l|s_j^l;\bm{\theta}^l) = \sigma(\bm{W}^l *\bm{s}_j^l + \bm{b}^l)
	\label{eq:action}
	\end{equation}
	
	\noindent where $\sigma(.)$ is the sigmoid function with the parameter $\bm{\theta}^l = \{\bm{W}^l, \bm{b}^l\}$.
	
	\noindent \textbf{Reward.}
	As there is no annotation on which words are related to a relation mention, we design a delayed reward to measure the adequacy of the extracted mention words once all the words in sentence $x_t$ are scanned. 
	The delayed reward consists of three terms: the word discriminability, the continuity of the relation mention, and the distance to the two entities.
	
	Formally, suppose a mention $m_t = w_{k_1}, w_{k_2}, \dots, w_{k_L}$ is extracted from sentence $x_t$, where $k_j$ ($1\leq j \leq L$) is a word index in $x_t$, and $L$ is the number of words in the extracted mention. We denote the indices of the two entities as $k_{e_1}$/$k_{e_2}$, respectively. 
	
	The delayed rewards is defined as:
	\begin{equation}
	\begin{aligned}
	r_{final}^l(x_t) & = \frac{ \mathcal{P}(r|x_t) - \mathcal{P}(r|x_t')}{\mathcal{P}(r|x_t)} \\ & - \lambda_1 \frac{k_L - k_1}{L} \\ &- \lambda_2 \frac{\sum_q|k_q - k_{e1}| + |k_q - k_{e2}|}{L} 
	\label{eq:reward_extractor}
	\end{aligned}
	\end{equation}
	\noindent where: \\
	1) The first term is the word discriminability which measures how well $m_t$ can distinguish the relation. $\mathcal{P}(r|x_t)$, defined by Eq. \ref{eq:reward_estimator} in the reward estimator, is the classification likelihood of sentence $x_t$. $x_t'$ is the sentence where $m_t$ are removed from $x_t$.
	\\
	2) The second term is 
	the continuity reward which encourages the extraction of a consecutive token span at a certain extent.
	\\
	3) The third term is 
	the distance reward which encourages that mention words should be close to the two entities.
	%

	The three rewards are soft constraints for mention extraction. For instance, the contituity reward encourages extraction of consective words, but the model may also extract non-consecutive words as mention.
	And, $\lambda_1/ \lambda_2$ are the hyper-parameters to balance the three factors.
	
	\subsection{Training Objective and Optimization}
	For the sentence selector, we aim to maximize the expected future cumulative rewards, as below:
	\begin{equation}
	\mathcal{J}(\bm{\theta}^h) = \mathbb{E}_{g_t\sim \mu(g_t^h|s_t^h;\bm{\theta}^h)} \left[R(g_t^h) \right]
	\end{equation}
	\noindent where $R(g_t^h)$ is the future cumulative rewards from state $s_t^h$. 
	To compute $R(g_t^h)$, we sampled some trajectories according to the current policy. Taking one trajectory $(s_1^h, g_1^h, \dots, s_n^h, g_n^h)$ as example ($n$ is the number of sentences in the top-level process), $R(g_t^h) = r_{final}^h + \sum_{k = t}^n \left[\gamma^{k - t} r_{final}^l(x_k) \right]$, Note that the rewards received by the low-level mention extractor $(r_{final}^l(x_k))$ are passed to the selector, which provides a feedback to indicate how well sentence selection is.
	
	Similarly, the mention extractor maximizes the expected cumulative rewards, as follows:
	\begin{equation}
	\mathcal{J}(\bm{\theta}^l) = \mathbb{E}_{a_t^l\sim \pi(a_t^l|s_t^l;\bm{\theta}^l)}\left[ R(g_t^l) \right]
	\label{eq:loss_extractor}
	\end{equation}
	\noindent where $R(g_t^l) = r_{final}^l$, since the mention extractor have no intermediate rewards but only a delayed final reward.
	
	According to the policy gradient theorem~\cite{PolicyGradient} and the REINFORCE algorithm~\cite{williams1992simple}, we compute the gradient of the top-level sentence selector policy as:
	\begin{equation}
	\begin{aligned}
	\nabla_{\bm{\theta}^h} \mathcal{J}(\bm{\theta}^h) = & \mathbb{E}_{g_t^h\sim \mu(g_t^h|s_t^h;\bm{\theta}^h)} \\ & \left[ R(g_t^h) \cdot \nabla_{\bm{\theta}^h} \log \mu(g_t^h|s_t^h;\bm{\theta}^h) \right]
	\end{aligned}
	\end{equation}
	
	The policy gradient of the low-level mention extractor yields:
	\begin{equation}
	\begin{aligned}
	\nabla_{\bm{\theta}^l} \mathcal{J}(\bm{\theta}^l) = & \mathbb{E}_{a_t^l\sim \pi(a_t^l|s_t^l;\bm{\theta}^l)} \\ & \left[ R(a_t^l) \cdot \nabla_{\bm{\theta}^l} \log \pi(a_t^l|s_t^l;\bm{\theta}^l) \right]
	\end{aligned}
	\end{equation}
	
	\begin{algorithm}[!t]
		\KwIn{Training data $\bm{T}$, and each relation $r$ has a sentence bag $X_r$.} 
		%
		\ForEach {$pair (r, X_r) \in \bm{T}$}
		{
			\ForEach {sentence $x_t \in X_r$}
			{
				Sample option $g_t^h$ for the selector:
				$g_t^h \sim \mu(g_t^h|s_t^h;\bm{\theta}^h)$, see Eq.~\ref{eq:option}; \\
				$r_t^h = 0$ ; \\
				\If {$g_t^h = 1$}{
					Sample actions for the extractor on sentence $x_t$with $\bm{\theta}^l$: \qquad $\{a_1^l, \dots, a_{m}^l\}, a_{j}\sim \pi(a_j^l|s_j^l )$, see Eq.~\ref{eq:action} ;\\
					Obtain the final reward from the extractor $r_{final}^l(x_t)$ ;\\
					Update the parameter $\bm{\theta}^l$ ;\\
					Compute the intermediate reward of the selector:
					$r_t^h = r_{final}^l(x_t)$, see Eq.\ref{eq:reward_extractor}
				}
				
			}
			Obtain the reward of the extractor $r_{final}^h$, see Eq.~\ref{eq:reward_selector} ;\\
			Update the parameter $\bm{\theta}^h$ ; \\
		}
		\caption{ Training Process of Hierarchical Reinforcement Learning}
		\label{alg:paper}
	\end{algorithm}
	
	
	For model learning, we first use all the sentences to pretrain a CNN classifier as the reward estimator and pretrain the low-level mention extractor according to Eq.~\ref{eq:cnn_loss} and Eq.~\ref{eq:loss_extractor} respectively.  
	After that, with the reward provided by the CNN classifier (parameters fixed), we are able to train the hierarchical RL model. See the details of our learning procedure in Algorithm~\ref{alg:paper}. 
	
	\subsection{Relation Mention Ranking}
	Note that our goal is to extract a set of representative phrases for a relation. Since our model extracts a mention from each selected sentence, we need to rank the extracted mentions at the corpus level to construct high-quality mention resources. 
	Formally, an extracted mention $m_i$ for a relation $r$ is ranked by the below score, similar to \cite{angeli2015leveraging}:
	\begin{equation}
	\mathcal{P}(m_i|r) \cdot \mathcal{P}(r|m_i)
	\label{eq-score}
	\end{equation}
	where $\mathcal{P}(m_i|r) = \frac{n(m_i, r)}{n(r)}$ and $\mathcal{P}(r|m_i) = \frac{n(m_i, r)}{n(m_i)}$. $n(m_i, r)$ is the times that mention $m_i$ is extracted for relation $r$, $n(r)$ is the number of the sentences labeled as relation $r$, and $n(m_i)$ is the times that mention $m_i$ is extracted from all the selected sentences.
	Finally, we select top $N$ mentions for each relation to construct the mention resource.
	
	\section{Experiments}
	\label{sec:exp}
	
	\subsection{Experimental Setup}
	\subsubsection{Data Preparation}
	We evaluated our model on a clean dataset and a noisy dataset, respectively.
	
    \vpara{Clean dataset.}
	The clean dataset is adopted from SemEval-2010~\cite{hendrickx2009semeval}, which contains 10,717 sentences and 9 distinct relations. The average sentence length is $20.0$. We took 8,000 sentences for training and the remainder for test. 
	
	\vpara{Noisy dataset.}
	To validate the performance of mention extraction from noisy data, we adopted a widely used
	dataset from~\cite{riedel2010modeling}\footnote{http://iesl.cs.umass.edu/riedel/ecml/}. 
	This dataset contains 522,611 sentences, 281,270 entity pairs, and 18,252 relational facts in the training set; 
	and 172,448 sentences, 96,678 entity pairs and 1,950 relational facts in the test set. 
	There are 39,528 unique entities and 53 unique relations. The average sentence length is $38.5$. This dataset consists of noisy sentences which may not describe a fact at all. 

		\begin{table}
		\centering
		\begin{tabular}{ c | c | c} \hline
			Method & Clean Data & Noisy Data \\ \hline
			StanfordIE & 0.30 & 0.11\\
			ATT & 0.27 & 0.02 \\
			N-gram & 0.38 & 0.24\\ \hline
			Single RL & \textbf{0.71} & 0.35 \\
			HRL & \textbf{0.71} & \textbf{0.52}\\ \hline
		\end{tabular}
		\caption{Sentence-level extraction accuracy for relation mention. Note tath HRL is the same as Single RL on the clean data. Note that StanfordIE is an unsupervised method.}
		\label{tb:mention}
	\end{table}
	
	\begin{table*}[h]
	    \centering
		\begin{tabular}{| l |} \hline
			\textit{Example-I:} the \textbf{Entity-Origin} relation between \textbf{name} and \textbf{address}. \\
			The headquarters of the operation were at Berlin and the code \textbf{[name]}$_{e_1}$ for the program was \textcolor{red}{derived from} that \textbf{[address]}$_{e_2}$. \\
			\textit{Output:} ATT: \underline{derived}$\big|$ StanfordIE: \underline{N/A}$\big|$ HRL: \underline{derived from}$\big|$\\ \hline 
			\textit{Example-II:} the \textbf{Product-Producer} relation between \textbf{philosopher} and \textbf{writings}. \\
			Andronicus wrote a work, the fifth book of which contained a complete list of the
			\textbf{[philosopher]}$_{e_1}$ \textcolor{red}{'s} \textbf{[writings]}$_{e_2}$. \\
			\textit{Output:} ATT: \underline{wrote}$\big|$ StanfordIE: \underline{of}$\big|$ HRL: \underline{'s} $\big|$\\ \hline
		\end{tabular}
		\caption{Examples for the extracted mentions by ATT, StanfordIE, and our model. N/A means StanfordIE did not extract any word.  }
		\label{tb:mention_case}
	\end{table*}
	
	\subsubsection{Baselines}
	
	\vpara{OpenIE}~\cite{angeli2015leveraging,mausam2012ollie}. OpenIE systems are the most relevant to our work, which extract a triple that contains two entity mentions and a relation mention. As aforementioned, OpenIE systems do not normalize different expressions to a canonical relation. Thus, we mapped the extracted mentions to a relation following the algorithm described in ~\cite{angeli2015leveraging}, which is trained on our training data. In our experiment, we use \textbf{Stanford OpenIE}~\cite{angeli2015leveraging} as baseline.
	
	\vpara{ATT}~\cite{huang2016attention}.
	ATT adopts a word-level attention over the words in a sentence and assigns each word an attention weight. We selected the word with the largest weight as the relation mention.
	
	%
	%
	
	\vpara{Single RL.} 
	This model only adopts the low-level mention extractor and ignores the top-level sentence selector. 
	We compared this model with our HRL model on the noisy dataset. On the clean dataset, HRL is unnecessary since there is no noisy sentences. 
	
	\vpara{N-gram.}
	To show the necessity for adopting reinforcement learning, we devised a new model named N-gram as our baseline, which searches over all n-grams ($n \le 3$) in a sentence and chooses as the mention the one which provides the maximal reward. The reward is the same as the final reward of the low-level mention extractor (see Eq. ~\ref{eq:reward_extractor}).

	\subsubsection{Parameter Settings}
	The parameters of our model are different on the clean and noisy datasets.
	For the clean dataset, we set the hyper-parameter $\lambda_1 = 1$, $\lambda_2 = 0.05$ and the learning rate as $0.01$. The training episode number is $50$.
	For the noisy dataset, we set $\lambda_1 = 0.4$, $\lambda_2 = 0.02$. The learning rate is $0.01$ and the training episode number is $5$ during the pretraining of the mention detector. The learning rate is $0.001$ and the training episode number is $50$ during the training of HRL.
	The reward discount factor is $\gamma=0.999$ on both datasets.
	
	For the parameters of the CNN classifier in the reward estimator, the word embedding dimension $d^w = 50$ and the position embedding dimension $d^p = 5$. The window size of the convolution layer $l$ is $3$. The learning rate is $\alpha = 0.02$. The batch size is fixed to $160$. The training episode number $L = 25$. We employed a dropout strategy with a probability of $0.5$.
	
	\subsection{Quality of Extracted Relation Mentions}
	We evaluated the the quality of extracted relation mentions with two metrics.
	At the sentence level, accuracy is assessed by manually checking whether the phrase extracted from a sentence is indeed representative for the given relation $r$.
	At the mention level, Precision@K is assessed by ranking the extracted mentions according to the representative ability (see Eq.~\ref{eq-score}). 
	

	\subsubsection{Sentence-level Evaluation}
	%
	We respectively sampled $300$ sentences from the clean and noisy datasets, and manually annotated the relation mention for each sentence. 
	As different baseline models extract multi-granularity relation mentions, we annotated  multiple relation mentions for each sentence for fair comparison. And, we guaranteed that all the annotations are representative for a given relation.
	For instance, for sentence ``Muscle fatigue is the number one cause of arm muscle pain.'' with relation label ``Cause-Effect'', mention  annotations are ``{\it is the number one cause of}'', ``{\it is the cause of}'', ``{\it the cause of}'' and ``{\it cause}''.
	
	Then, we compared the extracted mentions with those manual annotations for each sentence to evaluate the extraction performance.
	Thus, this is sentence-level evaluation. 
	The results shown in Table~\ref{tb:mention} reveal the following observations:
	
	\noindent \textbf{First}, our proposed models (Single RL and HRL) outperform the baselines on both clean and noisy data. 
	Compared to our model, ATT has two drawbacks: the word with the largest attention weight may not be a mention word; and it cannot identify a consecutive token span as mention. 
	As for StanfordIE, it failed to extract fact triples and did not extract any relation mention in many cases.
	
	\noindent \textbf{Second}, HRL outperforms the baselines substantially on the noisy data, demonstrating the effectiveness of data denoising by the sentence selector. 
	By contrast, StanfordIE, ATT, N-gram and Single RL all suffer from the noisy data remarkably due to the inability of excluding noisy sentences.
	
	We also note that ATT drops much more than other baselines on noisy data. Our investigation into the results shows that ATT is sensitive to the sentence length. The longer the sentence is, the more difficult ATT can locate the correct relation mention words. The average length of sentence in noisy data is much longer than that in the clean data ($38.5$ vs. $20.0$).
	
	\noindent \textbf{Third}, SingleRL outperforms N-gram on both clean and noisy data. The results show that our RL strategy  is reasonable and effective.

	We further presented some exemplar mentions extracted by the models in Table~\ref{tb:mention_case}.
	Interestingly, our model can not only identify typical phrases like ``{\it derived from}'', but also discover less typical representative words such as ``{\it 's'}'. For StanfordIE, it sometimes failed to extract any word or extracted undesirable results. As for ATT, it is prone to produce wrong attention. 
	
	\begin{table}[t]
		\centering
		\begin{tabular}{ c | c | c | c | c} \hline
			Method & P@1 & P@2 & P@5 & P@10 \\ \hline
			StanfordIE & 0.88 & 0.82 & 0.72 & 0.61 \\
			ATT & 0.67 & 0.74 & 0.61 & 0.44\\ 
			N-gram & 0.83 & 0.75 & 0.67 & 0.56 \\ \hline
			Single RL & \textbf{0.94} & \textbf{0.94} & \textbf{0.84} & \textbf{0.66} \\ \hline
		\end{tabular}
		\caption{Average Precision@K of the extracted mentions from the clean data (mention-level).}
		\label{tb:p@k:clean}
	\end{table}
	
	\begin{table}[t]
		\centering
		\begin{tabular}{ c | c | c | c | c} \hline
			Method & P@1 & P@2 & P@5 & P@10 \\ \hline
			StanfordIE & 0.38 & 0.50 & 0.40 & 0.33 \\
			ATT & 0.15 & 0.15 & 0.20 & 0.17\\
			N-gram & 0.38 & 0.38 & 0.42 & 0.37\\ \hline
			Single RL & 0.46 & 0.38 & 0.40 & 0.28 \\
			HRL & \textbf{0.77} & \textbf{0.77} & \textbf{0.74} & \textbf{0.71} \\ \hline
		\end{tabular}
		\caption{Average Precision@K of the relation mentions extracted from the noisy data (mention-level).}
		\label{tb:p@k:noisy}
	\end{table}
	
	\begin{table}[t]
		\centering
		{
			\renewcommand\arraystretch{1.2}
			\begin{tabular}{c|l}
				\hline
				\textbf{Relation} & \textbf{Exemplar phrases} \\
				\hline
				\multirow{2}*{Cause-Effect} & triggers, caused by, lead to, \\
				& generated by, instigates\\
				\hline
				\multirow{2}*{Product-Producer} & hand-made by, co-founded by, \\
				& makes, created by \\
				\hline 
				\multirow{2}*{Founder} & founder of, chief executive of, \\
				& managing director at, chairman of\\
				\hline
				\multirow{2}*{Children} & son of, daughter, \\
				& father, son of minister \\
				\hline 
			\end{tabular}
		}
		\caption{Exemplar mention phrases for some sampled relations.}
		\label{tb:lexicon}
	\end{table}
	
	\subsubsection{Mention-level Evaluation}
	We conducted mention-level evaluation to assess the quality of the extracted mentions at the corpus level. For each relation, we chose top 10 representative mentions which are ranked by Eq. \ref{eq-score}. 
	We adopted Precision@K as the performance metric.
	
	The results on the clean data and noisy data are presented in Table~\ref{tb:p@k:clean} and Table~\ref{tb:p@k:noisy}, respectively. 
	On the clean data, the top 5 mentions extracted by our model achieve a precision of more than 0.8, significantly higher than those obtained by StanfordIE and ATT (Table~\ref{tb:p@k:clean}). 
	As for the noisy data, P@10 drops remarkably for all the methods, but HRL performs much better than the baselines. 
	Moreover, HRL outperforms single RL remarkably . All the evidence supports that the sentence selector effectively exclude the noisy sentences (Table~\ref{tb:p@k:noisy}).  
	
	A concrete example of the ranked mentions is presented in Table~\ref{tb:lexicon}. It shows that the extracted phrases are representative and meaningful.
	We also show the top 10 relation mentions for some relations in a supplementary file.
	
	\subsection{Utility of Extracted Relation Mentions}
	We evaluated whether the extracted mentions can facilitate downstream applications such as relation classification, on both clean and noisy data. 

We first evaluated on the clean data how extracted mentions can benefit relation classification as addition feature. 
	More specifically, we constructed a binary vector where the $i$-th dimension represents whether at least one extracted mention of the $i$-th relation occurs in a given sentence. 
	The dimension of the binary vector equals to the number of relations. For each sentence, if it contains the mentions for a relation, the corresponding dimension will be set to $1$, otherwise $0$.
	
	\begin{table}
		\centering
		\begin{tabular}{ c | c | c } \hline
			Mention features & CNN & Regression\\ \hline
			StanfordIE & 81.74 & 27.52\\
			Ollie & 81.28 & 18.32\\
			ATT & 81.48 & 20.61\\ 
			N-gram & 81.57 & 36.97 \\ \hline
			Single RL & \textbf{82.13} & \textbf{39.32}\\ \hline
		\end{tabular}
		\caption{Macro $F_1$ of relation classification on the clean data.
		}
		\label{tb:re}
	\end{table}
	
    To fully check the effectiveness of the extracted relation mentions, we used the binary vector in two ways.
    The first way is that we directly used the binary vector for relation classification, with a logistic regression classifier. 
	The second way is to use the binary vector along with a CNN classifier. We concatenated the binary vector with the output of the pooling layer of a CNN structure, and fed the concatenated vector into a fully-connected layer for relation classification. 
	%
	
	We compared different mention features generated by Open IE, ATT, RL, and HRL respectively. The results on the clean data are shown in Table~\ref{tb:re}. It demonstrates that the relation mentions from our model obtain better performance than those from the baseline models. In the CNN classifier, mention features are only used as additional feature, which may explain that only slight improvement is observed.
	
	Due to the page limit, we provided a supplementary file to show the experiment results on noisy data. We believe that such extracted mentions would be beneficial for question answering, language inference, and more, which will be validated in future work.

	\subsection{Discussions}
	Our model is advantageous in extracting relation mentions that can be expressed explicitly by words. However, some relations are expressed implicitly, or sometimes, we need to make semantic reasoning to derive a relation.
	
	This first example demonstrates an implicit relation mention: sentence ``I spent a year working for a \textbf{[software]}$_{e_1}$ \textbf{[company]}$_{e_2}$ to pay off my college loans.'' is labeled with a {\bf Product-Producer} relation, which requires the knowledge that a software company sells software (However, the Apple company does not produce apple). 
	
	The second example shows that relation mention detection sometimes needs to make semantic reasoning: sentence
	``You 'll get an instant overview of \textbf{[Tallahassee]}$_{e_1}$, which was chosen as \textbf{[Florida]}$_{e_2}$ 's capital for only one reason $\cdots$'' is marked with the {\bf Contains} relation, which needs to be inferred from the capital relationship.
	The third example, ``\textbf{[Nicola Sturgeon]}$_{e_1}$, the newly elected first minister of \textbf{[Scotland]}$_{e_2}$, expressed concern that $\cdots$'', labeled with the {\bf Nationality} relation, also needs to make semantic reasoning from {\it minister} to derive the desired relation.

	Our model has limitations on these cases, and we will leave it as future work.
	
	\section{Conclusion}
	\label{sec:conclude}
	In this paper, we present a hierarchical reinforcement learning model for extracting relation mentions from noisy data. The model consists of a sentence selector to exclude noisy sentences, a mention extractor to identify mention words in a selected sentence, and a reward estimator to guide the policy learning of the selector and the extractor. 
	The model learns from large-scale noisy data without explicit annotations on either sentence (whether a sentence truly describes a relation) or on word (which words are a relation mention). 
	Experiments show that our model outperforms the state-of-the-art baselines. 
	

\bibliography{emnlp2018}

\begin{thebibliography}{}

\bibitem[\protect\citeauthoryear{Angeli, Premkumar, and
  Manning}{2015}]{angeli2015leveraging}
Angeli, G.; Premkumar, M. J.~J.; and Manning, C.~D.
\newblock 2015.
\newblock Leveraging linguistic structure for open domain information
  extraction.
\newblock In {\em ACL}, volume~1,  344--354.

\bibitem[\protect\citeauthoryear{Banko \bgroup et al\mbox.\egroup
  }{2007}]{banko2007open}
Banko, M.; Cafarella, M.~J.; Soderland, S.; Broadhead, M.; and Etzioni, O.
\newblock 2007.
\newblock Open information extraction from the web.
\newblock In {\em IJCAI}, volume~7,  2670--2676.

\bibitem[\protect\citeauthoryear{Borthwick \bgroup et al\mbox.\egroup
  }{1998}]{borthwick1998exploiting}
Borthwick, A.; Sterling, J.; Agichtein, E.; and Grishman, R.
\newblock 1998.
\newblock Exploiting diverse knowledge sources via maximum entropy in named
  entity recognition.
\newblock In {\em Sixth Workshop on Very Large Corpora}.

\bibitem[\protect\citeauthoryear{Bunescu and
  Mooney}{2005}]{bunescu2005shortest}
Bunescu, R.~C., and Mooney, R.~J.
\newblock 2005.
\newblock A shortest path dependency kernel for relation extraction.
\newblock In {\em EMNLP},  724--731.

\bibitem[\protect\citeauthoryear{Chiu and Nichols}{2016}]{chiu2016named}
Chiu, J.~P., and Nichols, E.
\newblock 2016.
\newblock Named entity recognition with bidirectional lstm-cnns.
\newblock {\em TACL} 4:357--370.

\bibitem[\protect\citeauthoryear{Dietterich}{2000}]{dietterich2000hierarchical}
Dietterich, T.~G.
\newblock 2000.
\newblock Hierarchical reinforcement learning with the maxq value function
  decomposition.
\newblock {\em J. Artif. Intell. Res.(JAIR)} 13(1):227--303.

\bibitem[\protect\citeauthoryear{dos Santos, Xiang, and
  Zhou}{2015}]{DBLP:conf/acl/SantosXZ15}
dos Santos, C.~N.; Xiang, B.; and Zhou, B.
\newblock 2015.
\newblock Classifying relations by ranking with convolutional neural networks.
\newblock In {\em ACL},  626--634.

\bibitem[\protect\citeauthoryear{Fader, Soderland, and
  Etzioni}{2011}]{fader2011identifying}
Fader, A.; Soderland, S.; and Etzioni, O.
\newblock 2011.
\newblock Identifying relations for open information extraction.
\newblock In {\em EMNLP},  1535--1545.

\bibitem[\protect\citeauthoryear{Feng \bgroup et al\mbox.\egroup
  }{2018}]{feng2018reinforcement}
Feng, J.; Huang, M.; Zhao, L.; Yang, Y.; and Zhu, X.
\newblock 2018.
\newblock Reinforcement learning for relation classification from noisy data.
\newblock {\em AAAI}.

\bibitem[\protect\citeauthoryear{Hendrickx \bgroup et al\mbox.\egroup
  }{2009}]{hendrickx2009semeval}
Hendrickx, I.; Kim, S.~N.; Kozareva, Z.; Nakov, P.; {\'O}~S{\'e}aghdha, D.;
  Pad{\'o}, S.; Pennacchiotti, M.; Romano, L.; and Szpakowicz, S.
\newblock 2009.
\newblock Semeval-2010 task 8: Multi-way classification of semantic relations
  between pairs of nominals.
\newblock In {\em Proceedings of the Workshop on Semantic Evaluations: Recent
  Achievements and Future Directions},  94--99.

\bibitem[\protect\citeauthoryear{Hoffmann \bgroup et al\mbox.\egroup
  }{2011}]{hoffmann2011knowledge}
Hoffmann, R.; Zhang, C.; Ling, X.; Zettlemoyer, L.; and Weld, D.~S.
\newblock 2011.
\newblock Knowledge-based weak supervision for information extraction of
  overlapping relations.
\newblock In {\em ACL},  541--550.

\bibitem[\protect\citeauthoryear{Huang and others}{2016}]{huang2016attention}
Huang, X., et~al.
\newblock 2016.
\newblock Attention-based convolutional neural network for semantic relation
  extraction.
\newblock In {\em COLING},  2526--2536.

\bibitem[\protect\citeauthoryear{Katiyar and Cardie}{2017}]{katiyar2017going}
Katiyar, A., and Cardie, C.
\newblock 2017.
\newblock Going out on a limb: Joint extraction of entity mentions and
  relations without dependency trees.
\newblock In {\em ACL}, volume~1,  917--928.

\bibitem[\protect\citeauthoryear{Li and Ji}{2014}]{li2014incremental}
Li, Q., and Ji, H.
\newblock 2014.
\newblock Incremental joint extraction of entity mentions and relations.
\newblock In {\em ACL}, volume~1,  402--412.

\bibitem[\protect\citeauthoryear{Lin \bgroup et al\mbox.\egroup
  }{2016}]{lin2016neural}
Lin, Y.; Shen, S.; Liu, Z.; Luan, H.; and Sun, M.
\newblock 2016.
\newblock Neural relation extraction with selective attention over instances.
\newblock In {\em ACL}, volume~1,  2124--2133.

\bibitem[\protect\citeauthoryear{Lin, Liu, and Sun}{2017}]{lin2017neural}
Lin, Y.; Liu, Z.; and Sun, M.
\newblock 2017.
\newblock Neural relation extraction with multi-lingual attention.
\newblock In {\em ACL}, volume~1,  34--43.

\bibitem[\protect\citeauthoryear{Mausam \bgroup et al\mbox.\egroup
  }{2012}]{mausam2012ollie}
Mausam; Michael, S.; Robert, B.; Stephen, S.; and Oren, E.
\newblock 2012.
\newblock Open language learning for information extraction.
\newblock In {\em EMNLP},  523--534.

\bibitem[\protect\citeauthoryear{Mintz \bgroup et al\mbox.\egroup
  }{2009}]{mintz2009distant}
Mintz, M.; Bills, S.; Snow, R.; and Jurafsky, D.
\newblock 2009.
\newblock Distant supervision for relation extraction without labeled data.
\newblock In {\em ACL-IJCNLP},  1003--1011.

\bibitem[\protect\citeauthoryear{Miwa and
  Bansal}{2016}]{miwa-bansal:2016:P16-1}
Miwa, M., and Bansal, M.
\newblock 2016.
\newblock End-to-end relation extraction using lstms on sequences and tree
  structures.
\newblock In {\em ACL},  1105--1116.

\bibitem[\protect\citeauthoryear{Mooney and
  Bunescu}{2005}]{mooney2005subsequence}
Mooney, R.~J., and Bunescu, R.~C.
\newblock 2005.
\newblock Subsequence kernels for relation extraction.
\newblock In {\em NIPS},  171--178.

\bibitem[\protect\citeauthoryear{Ren \bgroup et al\mbox.\egroup
  }{2017}]{ren2017cotype}
Ren, X.; Wu, Z.; He, W.; Qu, M.; Voss, C.~R.; Ji, H.; Abdelzaher, T.~F.; and
  Han, J.
\newblock 2017.
\newblock Cotype: Joint extraction of typed entities and relations with
  knowledge bases.
\newblock In {\em WWW},  1015--1024.

\bibitem[\protect\citeauthoryear{Riedel, Yao, and
  McCallum}{2010}]{riedel2010modeling}
Riedel, S.; Yao, L.; and McCallum, A.
\newblock 2010.
\newblock Modeling relations and their mentions without labeled text.
\newblock In {\em ECML-PKDD},  148--163.
\newblock Springer.

\bibitem[\protect\citeauthoryear{Socher \bgroup et al\mbox.\egroup
  }{2012}]{socher2012semantic}
Socher, R.; Huval, B.; Manning, C.~D.; and Ng, A.~Y.
\newblock 2012.
\newblock Semantic compositionality through recursive matrix-vector spaces.
\newblock In {\em EMNLP-CoNLL},  1201--1211.

\bibitem[\protect\citeauthoryear{Sutton \bgroup et al\mbox.\egroup
  }{1999}]{PolicyGradient}
Sutton, R.~S.; McAllester, D.; Singh, S.; and Mansour, Y.
\newblock 1999.
\newblock Policy gradient methods for reinforcement learning with function
  approximation.
\newblock In {\em NIPS}.

\bibitem[\protect\citeauthoryear{Sutton, Precup, and
  Singh}{1999}]{sutton1999between}
Sutton, R.~S.; Precup, D.; and Singh, S.
\newblock 1999.
\newblock Between mdps and semi-mdps: A framework for temporal abstraction in
  reinforcement learning.
\newblock {\em Artificial intelligence} 112(1-2):181--211.

\bibitem[\protect\citeauthoryear{Takamatsu, Sato, and
  Nakagawa}{2012}]{takamatsu2012reducing}
Takamatsu, S.; Sato, I.; and Nakagawa, H.
\newblock 2012.
\newblock Reducing wrong labels in distant supervision for relation extraction.
\newblock In {\em ACL},  721--729.
\newblock ACL.

\bibitem[\protect\citeauthoryear{Wang \bgroup et al\mbox.\egroup
  }{2016}]{wang2016relation}
Wang, L.; Cao, Z.; de~Melo, G.; and Liu, Z.
\newblock 2016.
\newblock Relation classification via multi-level attention cnns.
\newblock In {\em ACL}, volume~1,  1298--1307.

\bibitem[\protect\citeauthoryear{Williams}{1992}]{williams1992simple}
Williams, R.~J.
\newblock 1992.
\newblock Simple statistical gradient-following algorithms for connectionist
  reinforcement learning.
\newblock {\em Machine learning} 8(3-4):229--256.

\bibitem[\protect\citeauthoryear{Wu and Weld}{2010}]{wu2010open}
Wu, F., and Weld, D.~S.
\newblock 2010.
\newblock Open information extraction using wikipedia.
\newblock In {\em ACL},  118--127.

\bibitem[\protect\citeauthoryear{Xu, Jiang, and
  Watcharawittayakul}{2017}]{xu2017local}
Xu, M.; Jiang, H.; and Watcharawittayakul, S.
\newblock 2017.
\newblock A local detection approach for named entity recognition and mention
  detection.
\newblock In {\em ACL}, volume~1,  1237--1247.

\bibitem[\protect\citeauthoryear{Zeng \bgroup et al\mbox.\egroup
  }{2014}]{zeng2014relation}
Zeng, D.; Liu, K.; Lai, S.; Zhou, G.; and Zhao, J.
\newblock 2014.
\newblock Relation classification via convolutional deep neural network.
\newblock In {\em COLING},  2335--2344.

\bibitem[\protect\citeauthoryear{Zeng \bgroup et al\mbox.\egroup
  }{2015}]{zeng2015distant}
Zeng, D.; Liu, K.; Chen, Y.; and Zhao, J.
\newblock 2015.
\newblock Distant supervision for relation extraction via piecewise
  convolutional neural networks.
\newblock In {\em EMNLP},  1753--1762.

\bibitem[\protect\citeauthoryear{Zeng \bgroup et al\mbox.\egroup
  }{2018}]{zeng2018large}
Zeng, X.; He, S.; Liu, K.; and Zhao, J.
\newblock 2018.
\newblock Large scaled relation extraction with reinforcement learning.
\newblock {\em AAAI}.

\bibitem[\protect\citeauthoryear{Zheng \bgroup et al\mbox.\egroup
  }{2017}]{zheng2017joint}
Zheng, S.; Wang, F.; Bao, H.; Hao, Y.; Zhou, P.; and Xu, B.
\newblock 2017.
\newblock Joint extraction of entities and relations based on a novel tagging
  scheme.
\newblock In {\em ACL}, volume~1,  1227--1236.

\bibitem[\protect\citeauthoryear{Zhou \bgroup et al\mbox.\egroup
  }{2005}]{guodong2005exploring}
Zhou, G.; Su, J.; Zhang, J.; and Zhang, M.
\newblock 2005.
\newblock Exploring various knowledge in relation extraction.
\newblock In {\em ACL},  427--434.

\end{thebibliography}
\bibliographystyle{aaai}

\appendix
\section{Supplementary Materials}

	\begin{table*}[t]
	\centering
	{
		\renewcommand\arraystretch{1.2}
		\begin{tabular}{c|l}
			\hline
			\textbf{Relation} & \textbf{Top 10 relation mentions} \\
			\hline
			Cause-Effect & by, from, after, caused by, generated by, due, following, comes, through, produced by \\ \hline
			\multirow{2}*{Person-Company} & chief executive of, the, at, general, chairman of, president of, \\ & UNK cheif executive of, secretary general, vice president, founder of \\
			\hline
			Component-Whole & comprises, contains, has, includes, with, comprised, composed, a, consists of \\
			\hline
			Product-Producer & by, produced by, found by, created by, from, in, 's, secreted by, built by, from \\
			\hline
			Entity-Destination & into, to, sent to, in, placed into, migrated into, inside, placed, injected into, vested into \\
			\hline
		\end{tabular}
	}
	\caption{Top 10 relation mentions for some relations. 
	}
	\label{tb:lexicon}
\end{table*}

\subsection{Relation Mention Rankings}
We presented top 10 relation mentions for some relations in Table~\ref{tb:lexicon}. It shows that the extracted phrases are representative and meaningful.
Although most of the phrases are representative and meaningful, some of them lack semantic meanings, such as ``{\it by}'', ``{\it at}'' and ``{\it with}''.  To interpret these cases, we need to throw them back to the sentence context. For the 8th relation mention ``{\it UNK cheif executive of }'' for relation "Person-Company", ``UNK'' indicates the company name. 
\subsection{Utility of Extracted Relation Mentions}
	We evaluated whether the extracted mentions can facilitate downstream applications such as relation classification, on both clean and noisy data. The result on clean data is shown in the main paper. We show the result on noisy data in this supplementary file. 
	
	\subsubsection{Experiment on Noisy Data}
	Similar to the experiments on the clean data, we generated the binary vector for each sentence on the noisy data and concatenated it with the output of the pooling layer of a CNN, and fed the new vector into a fully-connected layer for relation classification. 
	
	As there is no manual annotation on noisy data, we evaluated the results under the held-out evaluation configuration, which provides an approximate measure of relation extraction without expensive human labors.

	We compared different mention features generated by HRL and the baseline models. We divided the baseline models into two groups. The first group consist of previous existing models include StanfordIE and ATT. The second group consists of the simplified version of HRL include SingleRL and N-gram.
	
\begin{figure}
		\centering
		\includegraphics[width=0.46\textwidth]{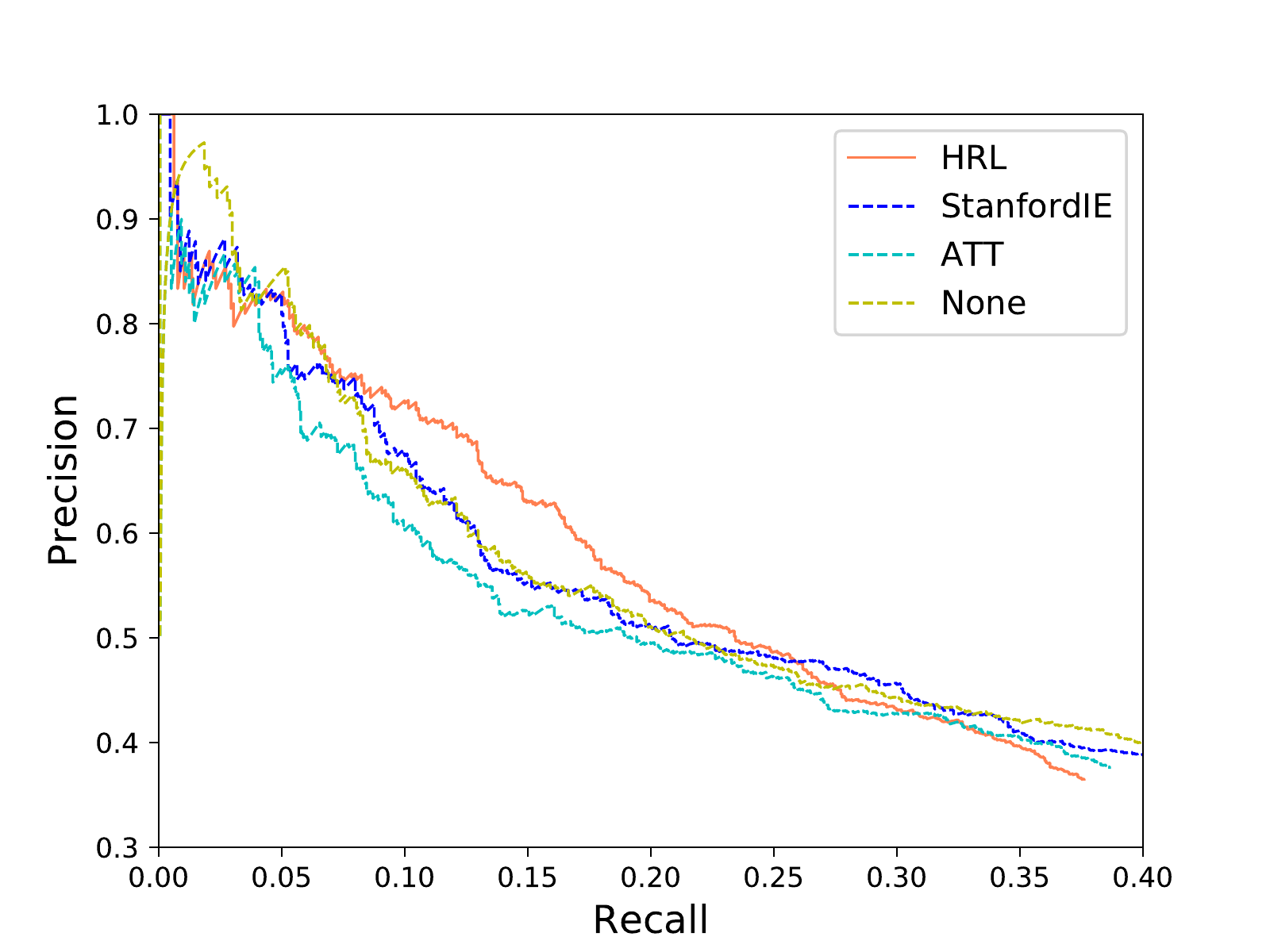}
		\caption{Comparison between HRL and the baselines. 
		}
		\label{fig:pr1}
	\end{figure}

\begin{figure}
		\centering
		\includegraphics[width=0.46\textwidth]{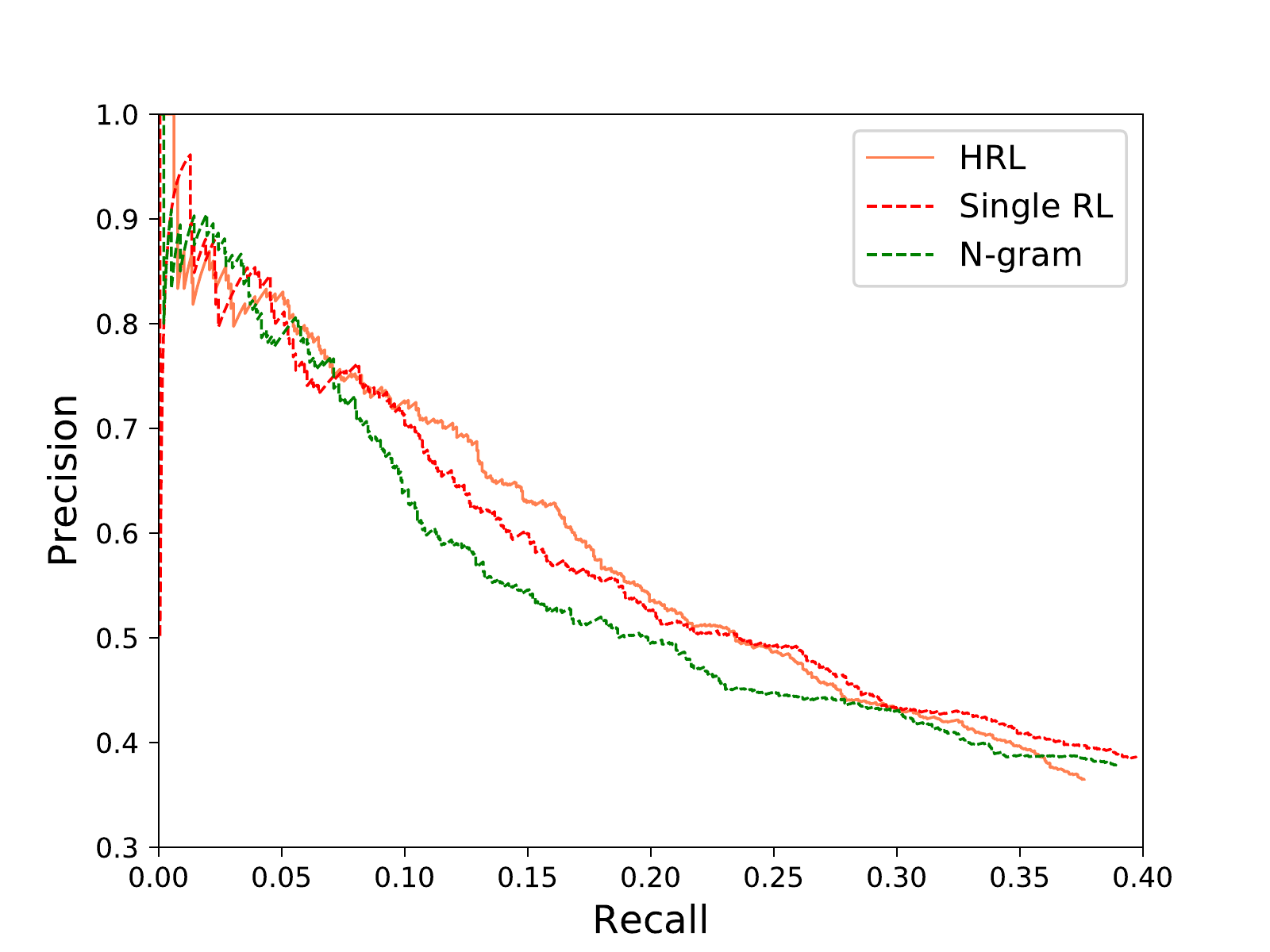}
		\caption{Comparison between HRL and its simplified models
		}
		\label{fig:pr2}
	\end{figure}

	Figure~\ref{fig:pr1} and Figure~\ref{fig:pr2} show the results on noisy data. Figure~\ref{fig:pr1} shows that our HRL model outperforms the existing mention extraction models. Figure~\ref{fig:pr2} shows that HRL outperforms Single RL and Single RL outperforms N-gram. This demonstrates that the necessity of removing noisy sentences for relation mention extraction.
	

\end{document}